\documentclass[journal]{IEEEtran}
\usepackage{amsmath,amsfonts}
\usepackage{algorithmic}
\usepackage{algorithm}
\usepackage{array}
\usepackage[caption=false,font=normalsize,labelfont=sf,textfont=sf]{subfig}
\usepackage{textcomp}
\usepackage{stfloats}
\usepackage{url}
\usepackage{verbatim}
\usepackage{graphicx}
\usepackage{cite}
\usepackage{xcolor}
\usepackage{colortbl}
\usepackage{nicematrix}

\begin{document}

\title{TelOps: AI-driven Operations and Maintenance for Telecommunication Networks}

\author{Yuqian Yang, Shusen Yang, Cong Zhao, Zongben Xu
\thanks{This work was supported in part by the National Key Research and Development Program of China under Grants 2021YFB2401300, 2022YFA1004100, and 2020YFA0713900; and in part by the National Natural Science Foundation of China under Grants 62172329, U1811461, U21A6005, and 11690011.}
\thanks{Y. Yang and C. Zhao are with the National Engineering Laboratory for Big Data Analytics, Xi’an Jiaotong University.}
\thanks{S. Yang and Z. Xu are with the National Engineering Laboratory for Big Data Analytics, and the Ministry of Education Key Lab for Intelligent Networks and Network Security, Xi’an Jiaotong University.}
\thanks{S. Yang is the corresponding author.}
}

\maketitle

\begin{abstract}
Telecommunication Networks (TNs) have become the most important infrastructure for data communications over the last century.
Operations and Maintenance (O\&M) is extremely important to ensure the availability, effectiveness, and efficiency of TN communications.
Different from the popular O\&M technique for IT systems (e.g., the cloud), Artificial Intelligence for IT Operations (AIOps), O\&M for TNs meets the following three fundamental challenges: topological dependence of network components, highly heterogeneous software, and restricted failure data.
This article presents TelOps, the first AI-driven O\&M framework for TNs, systematically enhanced with mechanism, data, and empirical knowledge.
We provide a comprehensive comparison between TelOps and AIOps, and conduct a proof-of-concept case study on a typical O\&M task (failure diagnosis) for a real industrial TN.
As the first systematic AI-driven O\&M framework for TNs, TelOps opens a new door to applying AI techniques to TN automation.
\end{abstract}


\section{Introduction}\label{sec:Introduction}
With the development of the Internet and communication techniques, Telecommunication Networks (TNs) become the most important Information and Communication Technology (ICT) infrastructure for data communications.
They have evolved into complex network systems comprising various wired (e.g., Metropolitan / Wide Area Networks) and wireless networks (e.g., 5G access networks).
Operations and Maintenance (O\&M) for TNs is vital to guarantee the effective and efficient running of complex, hybrid, and large-scale TNs~\cite{tmf2021an}.

O\&M for TNs has become an increasingly challenging task for all telecommunication industries~\cite{ig1218}.
Traditional O\&M heavily relies on expert experience and human labor, and is incapable for fast-developing modern TNs~\cite{ig1193}.
TM Forum, the leading association of telecommunication industries, recently depicts future TNs as \textit{\textbf{Autonomous Networks}}~\cite{tmf2021an}, where technologies like Artificial Intelligence (AI) need to be introduced for the automation, self-healing, and self-optimization of TNs.
AI-driven O\&M is indispensable to autonomous TNs, but there still lacks a systematic framework.
Different from the popular O\&M technique for IT systems (e.g., the cloud), Artificial Intelligence for IT Operations (AIOps)~\cite{dang2019aiops}, AI-driven O\&M for TNs meets the following three fundamental challenges:

\begin{enumerate}
\item \textbf{Topological Dependence of Network Components:}
TNs are distributed in nature.
Its O\&M usually covers all network devices (e.g., routers, switches) and communication links (e.g., optical fibers, wireless links) in a network system (e.g., a 5G network).
Components of TNs are more dependent on the network topology~\cite{fulber2020network} than typical IT systems (e.g., a cloud datacenter).

\item \textbf{Highly Heterogeneous Software:}
Hybrid TNs comprise heterogeneous devices from different vendors and installed with vendor-specific software.
Additionally, both the costly-thus-passive unified upgrade and the continuous addition of new devices in existing TNs further exacerbate the device software heterogeneity.
These induce multi-source runtime information from heterogeneous devices across the network~\cite{wang2019tensor}.
Such a fact is fundamentally different from typical IT systems.

\item \textbf{Restricted Failure Data:}
As vital infrastructures, TNs are usually immediately restored at any cost to prevent the re-occurrence of the same type of failures.
It is almost impossible to reproduce all the services and states at the TN failure time for explicit studies.
Such a fact further reduces the inherently scarce data of each type of TN failure.
For IT systems, differently, failure data are usually sufficient.
\end{enumerate}

This article presents TelOps, the first systematic O\&M framework for TNs empowered by knowledge-enhanced AI.
Particularly, TelOps classifies essential O\&M functions for TNs into different layers, from the physical layer managing both the target TN and O\&M experts, to the application layer hosting various preventive and reactive O\&M tasks.
A comprehensive comparison between TelOps and AIOps is also provided.
We conduct a proof-of-concept case study on the failure diagnosis for a real industrial Mobile Access Network (MAN).
Results demonstrate that TelOps achieves significant performance gains (up to 28.0\% higher diagnosis accuracy, and better generalization capability) by introducing mechanism, data, and empirical knowledge of TNs into general machine learning methods.
We also discuss important open research issues of AI-driven O\&M for TNs.

\section{The Layering Architecture of TelOps}\label{sec:Overview}
\begin{figure*}[ht]
	\centering
    \includegraphics[width=0.95\textwidth]{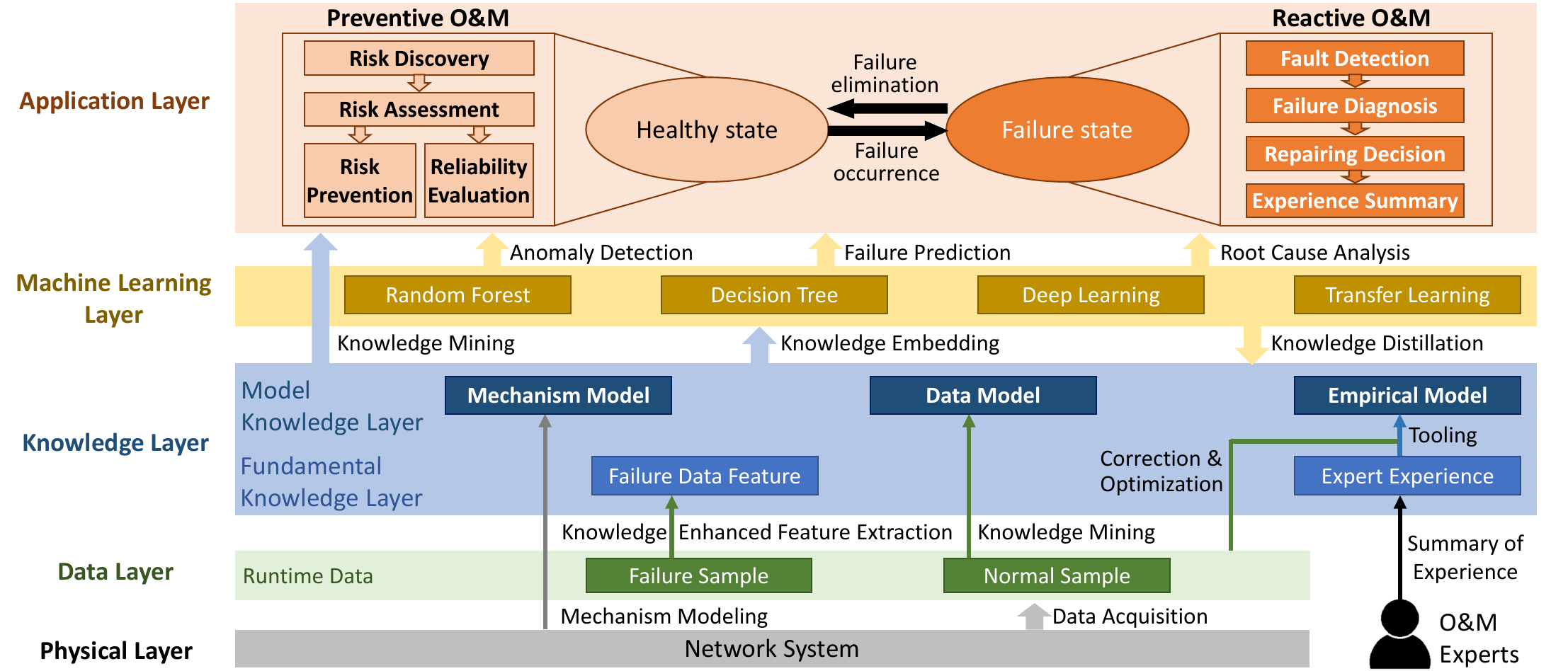}
    \caption{The layering architecture of TelOps.}
	\label{fig:sys-arch}
\end{figure*}

TelOps adopts a layering architecture comprising the \textit{\textbf{application layer}} hosting various O\&M tasks for TNs, and the \textit{\textbf{machine learning}}, \textit{\textbf{knowledge}}, \textit{\textbf{data}}, and \textit{\textbf{physical}} layers providing systematic support to O\&M tasks at the application layer.
The architecture of TelOps is shown in Fig.\ref{fig:sys-arch}.

\textit{\textbf{Application Layer:}}
This hosts various O\&M tasks for TNs.
Considering whether the system is currently healthy or not, these tasks have different goals.
When the system is healthy, preventive O\&M tasks (from risk discovery to reliability evaluation) avoid potential system failures.
When a failure occurs, reactive O\&M tasks (from fault detection to experience summary) focus on the elimination of the failure.
All under layers provide systematic support to application layer tasks.

\textit{\textbf{Machine Learning Layer:}}
This focuses on designing application-specific machine learning methods for various O\&M tasks.
The key principle is to subtly integrate proper O\&M knowledge into general machine learning algorithms, e.g., Convolutional Neural Networks (CNNs), Long Short-Term Memory (LSTM) and Graph Neural Networks (GNNs), for better O\&M performance, e.g., fault detection accuracy.

\textit{\textbf{Knowledge Layer:}}
This extracts and manages various knowledge contributing to O\&M tasks.
Unlike AIOps, empirical knowledge extracted from expert experience is critical but insufficient to O\&M for TNs.
Due to the ever-increasing complexity of TNs, expert experience is usually local and fragmented.
Therefore, mechanism knowledge from TN's intrinsic laws (e.g., topology, communication protocols) and data knowledge underneath runtime information are indispensable to the selection and optimization of the upper layer algorithms.

Particularly, such knowledge is further divided into fundamental and model knowledge sub-layers.
Fundamental knowledge (e.g., failure data features) is usually directly derived from system runtime data and O\&M experts.
It can be reused to construct different task-specific model knowledge (e.g., the fault propagation model) as the input of the upper layer machine learning algorithms.

\textit{\textbf{Data Layer:}}
This organizes TN runtime raw data (e.g., device logs, performance traces) collected from key hardware and software components in different network devices.
Both TN failure and normal samples are extracted for data knowledge mining, or as the input of machine learning algorithms.

\textit{\textbf{Physical Layer:}}
This manages TNs as network systems with Human-in-the-Loop.
All O\&M knowledge for TNs comes from two kinds of entities, the target TN itself and the O\&M experts.
Experts are indispensable since their experience contributes as essential fundamental knowledge.

\section{TelOps Tasks at the Application Layer}\label{sec:tasks}
We classify TelOps tasks into two categories before and after the occurrence of TN failures: \textit{\textbf{preventive}} and \textit{\textbf{reactive}} tasks.
We first define the failure, fault, and risk of TNs.

\begin{itemize}
\item \textbf{Failure} is the state that the target TN is not running as expected.
It may lead to system accidents.
There are different failures at different system scales.
For example, severe interference from neighboring districts causes failures on cell sites in one particular district.
The core network server offline failure regards the entire network.

\item \textbf{Fault} is the state that one single component malfunctions.
For any component, its faults are a subset of all its abnormal states.
For example, an excessive single signal of Multiple-Input Multiple-Output (MIMO) antennas on the base station side is an abnormal state, but an excessive overall signal is a fault.

\item \textbf{Risk} indicates the probability of failure occurrence.
For example, a bus topology network without any backup link has a high risk of network disconnection, since the failure of any node between the terminal node and the core network will cause a disconnection failure.
\end{itemize}

\begin{figure*}
	\centering
	\includegraphics[width=0.95\linewidth]{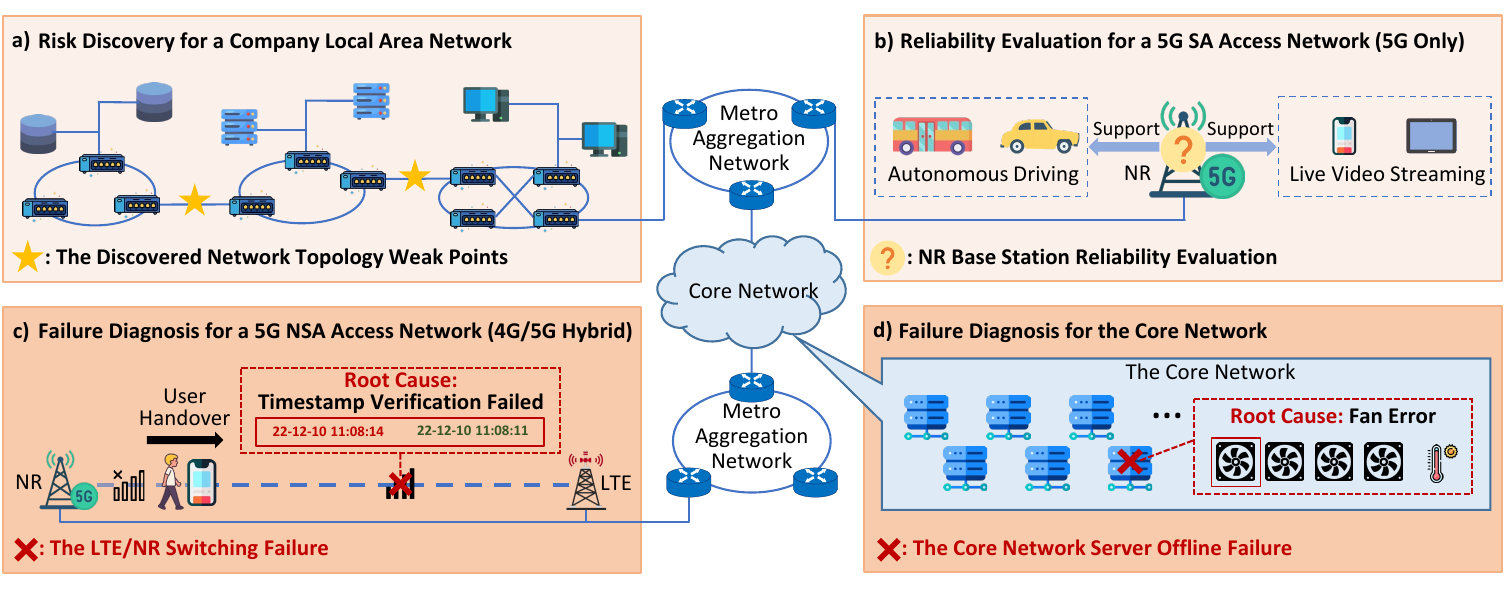}
 \caption{Representative Preventive and Reactive O\&M Tasks for a Typical TN.}
	\label{fig:tn-tasks}
\end{figure*}

Several representative preventive and reactive O\&M tasks for a typical TN are illustrated in Fig.\ref{fig:tn-tasks}.

\subsection{Preventive TelOps Tasks}
Preventive TelOps tasks aim at the maintenance of TNs before failures happen, i.e., dealing with risks.
Essential preventive tasks include risk discovery, risk assessment, risk prevention, and reliability evaluation.

\begin{itemize}
\item \textbf{Risk Discovery} identifies risks in the current system.
Lacking a systematic model for all TN failures, there exist bare risk discovery solutions currently.
The understanding of networking mechanisms~\cite{furdek2021optical} and expert experience generally helps to identify TN weak points.
For the local area network in Fig.\ref{fig:tn-tasks}, according to the graph theory, all links in the cut of its network topology are weak points.

\item \textbf{Risk Assessment} identifies the existence of risks in TNs, providing a risk indicator to help O\&M experts determine whether to intervene~\cite{furdek2021optical}.
Currently, academia usually relies on failure prediction~\cite{huang2019proactive}, while the industry often adopts fault detection from other TNs, which may not be effective due to the lack of isolation of TN-specific properties.
Transfer learning is promising to empower experiences of existing fault detection solutions.

\item \textbf{Risk Prevention} determines actions to take to prevent failures.
The major challenge is the automation and optimization of existing human-based actions~\cite{tmf2021an}.
Shared O\&M actions can be taken to prevent different risks, indicating that latent protocols can be mined to optimize specific actions on certain TN components (e.g., time synchronization to avoid timestamp mismatch).

\item \textbf{Reliability Evaluation} derives system reliability criteria from component runtime information and configuration~\cite{kafi2017survey}.
For the 5G SA access network in Fig.\ref{fig:tn-tasks}, evaluating the service reliability of the NR base station is crucial for supporting emerging killer applications like autonomous driving and live video streaming.
Lacking the fundamental understanding of complex TNs, there exists no systematic solution currently.
In fact, such criteria should be constructed hierarchically, i.e., from component-wise to TN topology-aware system-wise reliability evaluation.
\end{itemize}

\subsection{Reactive TelOps Tasks}
Reactive TelOps tasks aim at repairing system failures.
Essential reactive tasks include fault detection, failure diagnosis, repairing decision, and experience summary.

\begin{itemize}
\item \textbf{Fault Detection} is to identify device or service faults early on to prevent cascading failures.
System failure is usually caused by the fault of one or more components, where a certain indicator is often abnormal.
The fault detector needs to determine whether the abnormal component is faulty.
Popular methods using AIOps are usually data-intensive (e.g., Auto-Encoder~\cite{huang2020machine}).
For TNs with restricted failure data, it is essential to use methods that incorporate operational knowledge and runtime data for accurate fault detection.

\item \textbf{Failure Diagnosis} identifies the failure root cause (generally specific component faults) by analyzing numerous system representations (e.g., warnings, alerts).
For the 5G NSA access network in Fig.\ref{fig:tn-tasks}, when there occurs the LTE/NR switching failure during user handover, the root cause like `timestamp verification failed' needs to be rapidly identified according to traces and logs of the base station.
Similarly, in the core network, when a certain server is offline, the root cause like `fan error' must be located based on service logs.
Root Cause Analysis (RCA) is the predominating method aiming at directly finding the root cause.
Its effectiveness, however, is severely restricted in complex TNs.
Two-step methods~\cite{yang2019efficient} that identify the root component first, then the root cause, have limited effectiveness as they do not consider the inter-component fault propagation.
A more effective solution is to directly embed empirical knowledge in RCA.

\item \textbf{Repairing Decision} determines optimal repairing action sequences to restore the system.
Existing industrial solutions are manual and non-optimal, posing challenges in decision optimization and automation~\cite{tmf2021an}.
Action sequences are temporally dependent (e.g., counter value extraction should be conducted before counter reset), and topology-specific mechanism knowledge could improve decision optimization by leveraging the directed graph and network topology correlation.

\item \textbf{Experience Summary} sums up failure diagnosis and repairing experiences, aiming at compiling key knowledge to guide future reactive tasks.
Industrial solutions rely on formalized documents, lacking the condensed knowledge needed for intelligent O\&M.
The neural network is a promising knowledge model, which can be used to efficiently embed different types of knowledge.

\end{itemize}

\section{Supporting Layers of TelOps}\label{sec:support}

\begin{table*}[!t]
	\renewcommand{\arraystretch}{1.1}
	\caption{Desired machine learning methods and specific knowledge to embed for different O\&M tasks of TNs}
	\label{tab:pref}
	\begin{NiceTabular}{llXl}[hvlines]
        \CodeBefore
            \rowcolor{orange!60}{1}
            \rowcolor{orange!5}{2-5}
            \rowcolor{orange!15}{6-9}
        \Body
    	\textbf{Category} & \textbf{Task} & \textbf{Desired Machine Learning Methods} & \textbf{Knowledge to Embed}\\
		\Block{4-1}{Preventive Tasks} & Risk Discovery & Multi-Layer Perceptron (MLP), Clustering~\cite{yang2022massive} & Risk Feature, Network Topology\\
	    & Risk Assessment & LSTM, Gate Recurrent Unit (GRU), SVM & Risk Feature, Network Topology\\
		& Risk Prevention & Logistic Regression, Reinforcement Learning & System Operational Knowledge, Risk Feature\\
        & Reliability Evaluation & MLP, CNN, Random Forest, Decision Tree & System Operational Knowledge, Network Topology\\
		\Block{4-1}{Reactive Tasks} & Fault Detection & MLP, CNN, LSTM, GRU, Auto-Encoder~\cite{huang2020machine} & Component Operational Knowledge, Fault Feature\\
		& Failure diagnosis & CNN, LSTM, GRU, GNN~\cite{pujol2021ignnition}, Random Forest & System Operational Knowledge, Failure Feature\\
		& Repairing Decision & Random Forest, Reinforcement Learning & System Operational Knowledge, Network Topology\\
        & Experience Summary & MLP, CNN, GNN, Support Vector Machine & System Operational Knowledge, Failure Feature\\
	\end{NiceTabular}
\end{table*}

To fulfill the O\&M tasks at the application layer, we propose four functional layers (from the physical to machine learning layers) to provide systematic support.

\textit{\textbf{Physical Layer:}}
For the target TN, runtime information is acquired and digitalized, and the generated raw data are passed to the data layer.
Knowledge of system composition and running is directly gained via mechanism modeling (e.g., traditional modeling via experts' understanding of TNs).
Knowledge of O\&M experts is summarized as expert experience.

\textit{\textbf{Data Layer:}}
Due to the huge data size generated in real-time, modern TNs only keep selected information as logs, traces, time-series data, and raw binary data.
Data are stored distributively since only limited bandwidth is reserved for O\&M.
With data-driven methods, knowledge of data correlations, trends, and characteristics is mined directly from this layer.
Extracted failure data features serve different O\&M tasks at the application layer.
Runtime data are also used to optimize tools extracting empirical models from expert experience.

\textit{\textbf{Knowledge Layer:}}
This is the core of TelOps.
The fundamental knowledge layer contains preliminarily processed knowledge including failure data features, failure text description features, expert experience, etc.
The model knowledge layer contains three different types of primitive models: the mechanism, data, and empirical models, selectively organized as task-oriented representations for the machine learning layer.
Particularly, the mechanism model, e.g., the data rate formula in Subsection 4.1.2 of 3GPP-TS38.306~\cite{ts38306}, represents the intrinsic laws and patterns of the target TN and its failures, offering insights for designing the neural network structure and model update scheme.
The data model, e.g., a Directed Acyclic Graph (DAG) implying device associations, represents features, trends, and associations directly learned from data, providing valuable input for refining task-oriented features.
The empirical model, e.g., a fault tree representing organized expert rules for failure diagnosis, contains all executable rules observed and abstracted by O\&M experts, enhancing the interpretability of the overall solution through explainable empirical local models.

\textit{\textbf{Machine Learning Layer:}}
Typical general machine learning algorithms used by O\&M include anomaly detection, failure prediction, RCA, knowledge distillation, etc.
Such algorithms are enhanced by various knowledge from under layers to construct task-specified machine learning methods for O\&M of TNs.
For instance, \cite{yang2019efficient} uses the network topology to enhance a Hopfield neural network in failure diagnosis.
\cite{lv2021deep} demonstrates that deep neural networks with mechanism knowledge like beamforming can address failure diagnosis and repairing decision in 5G networks.

We summarize desired machine learning methods and knowledge to embed for different TN O\&M tasks in Table~\ref{tab:pref}.

\section{TelOps vs. AIOps}\label{sec:comparison}
A core difference between TelOps and AIOps is that, unlike AIOps, different types of knowledge need to be embedded in TelOps considering TNs' inherent properties.
Here, we provide a comprehensive comparison between TelOps and AIOps.

\subsection{Comparison of the Physical Layer}
O\&M experts are indispensable in TelOps since their experience plays a vital role in practice, but optional for AIOps.
In IT systems (e.g., the cloud), data-driven approaches can replace expert experience due to abundant training data and labels. It is not feasible for TNs with restricted failure data.

\subsection{Comparison of the Data Layer}
TelOps suffers from poor data quality (e.g., inconsistent sampling rates) due to system heterogeneity, unlike AIOps that can leverage abundant public data with established standards.
Restricted data for each type of TN failure also prevents the direct application of data-intensive methods like generative adversarial networks for data augmentation.

\subsection{Comparison of the Knowledge Layer}
Knowledge is an indispensable part of TelOps, but is not mandatory in AIOps.
O\&M tasks for over-complex TNs usually need guidance from expert experience and TN mechanism knowledge.
AIOps, however, could directly use general machine learning algorithms.
Additionally, since TN knowledge is usually long-term effective (once deployed, there are few chances for a TN to be adjusted in the recent several years), it is rational to treat the individual knowledge layer as the core of the AI-driven O\&M framework for TNs.
Knowledge in AIOps, differently, is updated much more frequently due to the rapidly evolving system software and hardware.

\begin{table}[!t]
	\renewcommand{\arraystretch}{1.1}
	\caption{Differences between TelOps and AIOps}
	\label{tab:diff}
	\centering
    \begin{NiceTabular}{X[6,l,m]X[14,l,m]X[8,l,m]X[7,l,m]}[hvlines]
        \CodeBefore
            \rectanglecolor{orange!60}{1-1}{6-2}
            \rectanglecolor{orange!60}{1-3}{1-4}
            \rectanglecolor{blue!15}{2-3}{6-3}
            \rectanglecolor{yellow!15}{2-4}{6-4}
            \rectanglecolor{orange!15}{2-1}{6-2}
        \Body
	    \textbf{Layer} & \textbf{Aspects} & \textbf{TelOps} & \textbf{AIOps}\\
        \textbf{Physical Layer} & Managed Entities & Target TN, O\&M Experts & Target IT System\\
		\textbf{Data Layer} & Data Quality & Poor & Fair\\
		\textbf{Knowledge Layer}& Knowledge Requirement & Mandatory & Optional\\
		\Block{2-1}{\textbf{Machine Learning Layer}} & Algorithm Selection & Knowledge guided & Data based/ Non-guided\\
        & Method Construction & Knowledge guided & Non-guided\\
    \end{NiceTabular}
\end{table}

\subsection{Comparison of the Machine Learning Layer}
Machine learning methods in TelOps are specifically designed with knowledge guidance. 
TelOps retrofits general machine learning algorithms with task-specified knowledge.
Differently, AIOps usually adopts general machine learning methods only based on the data format, or even with no reason.
For failure diagnosis, TelOps selects the GNN considering the TN fault propagation process.
In AIOps, LSTM is often intuitively selected for text processing.

We summarize the differences between TelOps and AIOps at different layers in Table~\ref{tab:diff}.

\begin{figure*}[!t]
	\centering
	\includegraphics[width=0.8\linewidth]{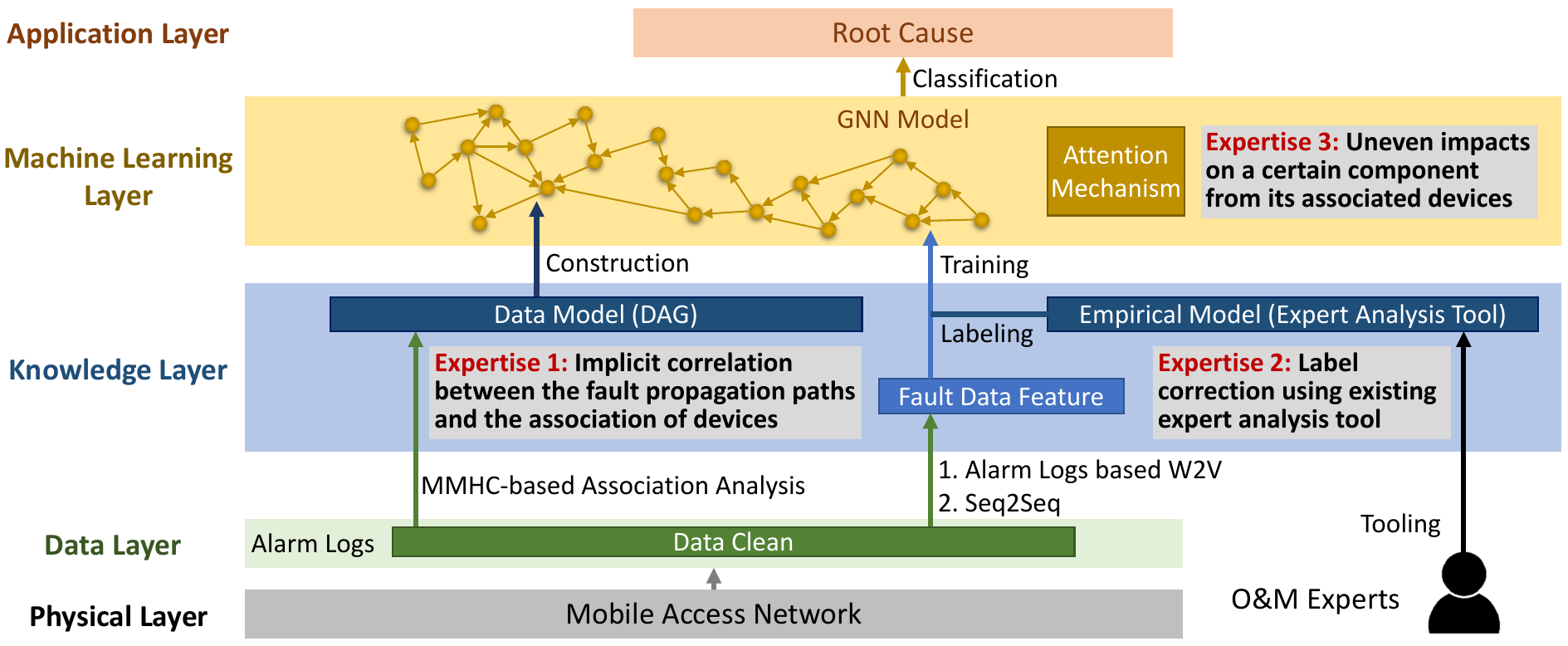}
	\caption{The workflow of failure diagnosis for MANs using TelOps.}
	\label{fig:case}
\end{figure*}

\section{Case study: Failure Diagnosis for Mobile Access Networks with TelOps}\label{sec:case}
To demonstrate the effectiveness of TelOps, we conduct a proof-of-concept case study on the failure diagnosis for a real industrial Mobile Access Network (MAN).
Failure diagnosis is extremely important since inaccurate diagnosis leads to not only untimely system reparation but also fatal system errors.
It is the most common O\&M task for industrial TNs: according to the statistics from our partner, a leading global provider of ICT infrastructure and smart devices, more than half of O\&M tasks for TNs are failure diagnosis tasks.

\subsection{The Failure Diagnosis Task for MANs}
Modern MANs comprise numerous system components collaborating intensely.
Once a fault occurred at one device, it would quickly trigger alarms at all associated devices, generating a flood of alarms.
Alarm information is mostly generated and stored as logs.
The goal of our failure diagnosis task is to reason out the root cause according to all alarm information after the failure occurs.

\subsection{Failure Diagnosis for MANs using TelOps}
The fault propagation pattern is core to determining the root cause of TN failures.
It cannot be explicitly modeled due to the complexity of modern MANs.
Experienced experts in the industry usually use the result of device association analysis as an alternative.
However, the latent fault propagation features of MANs cannot be effectively captured by predominating association analysis algorithms like Apriori and Max-Min Hill-Climbing (MMHC).
With TelOps, guided by the above expert experience, we constructed a GNN-based failure diagnosis method for MANs, whose workflow is illustrated in Fig.\ref{fig:case}:

\begin{enumerate}
\item \textbf{Data Layer:}
We cleaned alarm logs collected from all components in the target MAN, eliminating entries with missing key values (e.g., alarm name).

\item \textbf{Knowledge Layer:}
Considering the expertise that fault propagation in TNs is implicitly correlated with device associations, we constructed a DAG to extract device associations, supporting more precise propagation path capturing at the machine learning layer.
To address heterogeneous logs, we formed a unified fault data space using W2V and seq2seq models.
We also corrected a part of fault labels using our partner's empirical expert analysis tool.

\item \textbf{Machine Learning Layer:}
We created a GNN to learn the critical latent fault propagation features of MANs. 
The architecture of GNN was derived from the knowledge layer, where all edges were inherited from the DAG, and all vertices were represented using the unified fault data space.
Most importantly, considering the expertise that different associated components have uneven impacts on a certain component, we integrated an attention layer into the GNN.
The GNN was trained with all labeled samples enhanced at the knowledge layer.
For inference, given all alarm logs, the GNN outputs the failure root cause.
\end{enumerate}

\subsection{Experimental Setups}
Extensive experiments using alarm logs from a real industrial MAN were conducted to validate our method.
\begin{itemize}
\item
\textbf{Implementation:}
We used Pytorch (v1.3.0), an open-source machine learning framework, to implement our method.
All evaluations were conducted on a Linux server (64-bit Ubuntu 18.04, Linux kernel 4.15.0) with an Intel Xeon Gold 6130 CPU and 512GB RAM.
\item
\textbf{Dataset:}
We used a real-world dataset from our partner for validation.
It contains raw alarm data collected from the MAN of a South-East Asian country within three months.
There are 5.30 million alarm records in total, and each record has 193 columns.
\item
\textbf{Preprocessing:}
For each failure root cause, we identified its corresponding root alarm record with expert experience.
For each identified record, we extracted all records collected within five minutes before and after the record.
Tuples containing extracted records and the failure root cause were treated as TelOps' input.
Preprocessing, including data loading, took 10 hours.
\item
\textbf{Comparatives:}
We compared our approach with two conventional industrial methods, Random Forest and CNN, and an AIOps method.
Random Forest is currently used by our partner due to its high interpretability and adjustable weights in expert experience aggregation.
CNN is the best-performing method using neural networks investigated by our partner.
With no existing method, we constructed an intuitive AIOps solution following \cite{pujol2021ignnition}, which generated a GNN sharing the same fault feature vector as TelOps but using a fully connected graph.
\end{itemize}

\subsection{Results}
\begin{figure}[!t]
	\centering
    \includegraphics[width=0.488\textwidth]{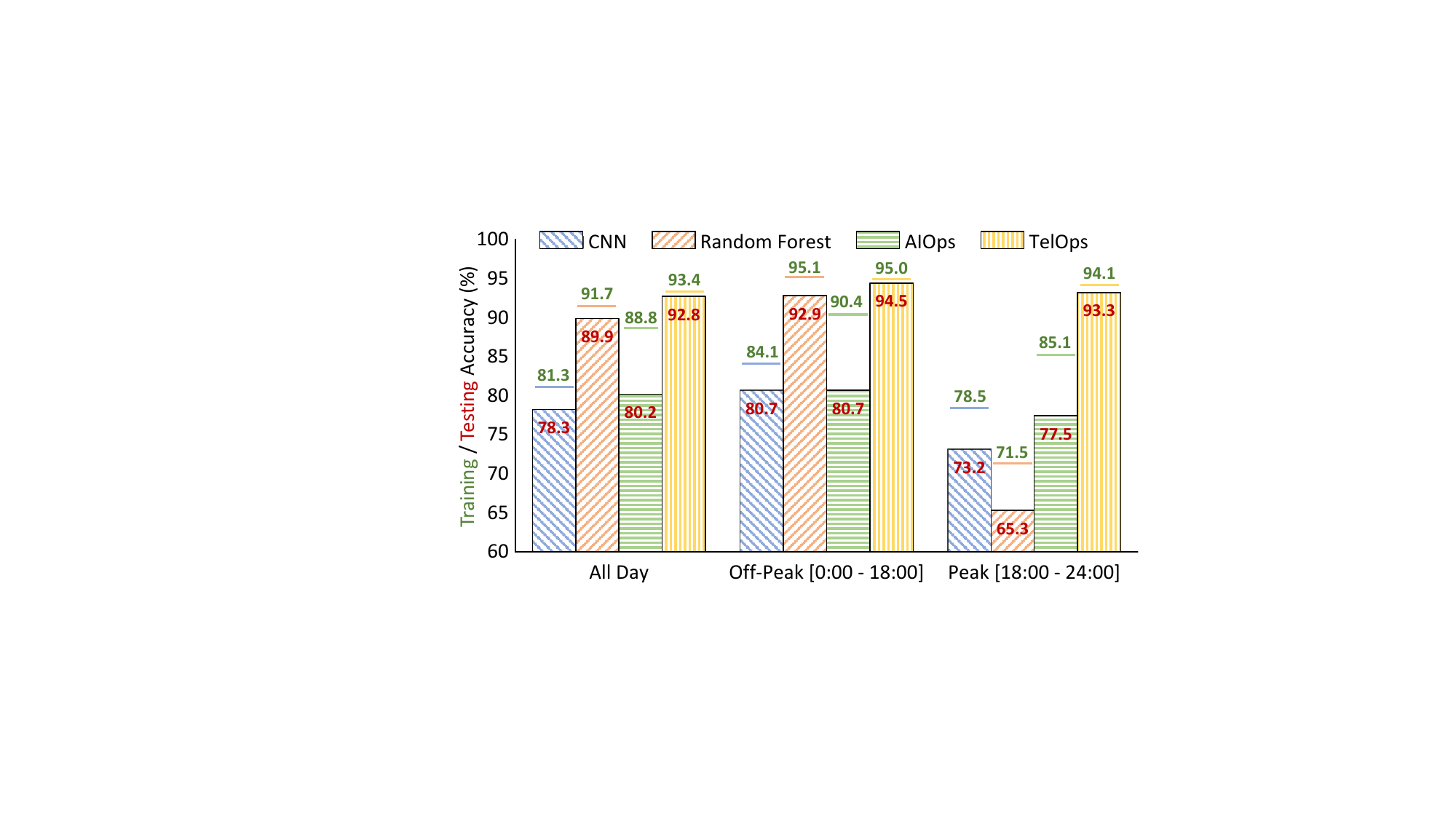}
    \caption{The accuracy of failure root cause identification for the MAN.}
	\label{fig:exp}
\end{figure}

We conducted extensive performance evaluation under three TN operating scenarios, i.e., `All Day', `Off-Peak [0:00-18:00]' covering bedtime and office hours, and `Peak [18:00-24:00]' covering after-work leisure time, containing 5.30, 3.96, and 1.34 million  alarm records, respectively.
Model training on the `All Day' dataset took 37 minutes for TelOps and AIOps, and 22 minutes for CNN.
Fig.\ref{fig:exp} illustrates the failure diagnosis accuracy (i.e., testing accuracy) of all methods.

According to Fig.\ref{fig:exp}, TelOps achieves the highest diagnosis accuracy (92.8\%$\sim$94.5\%) among all comparatives, and maintains a prominent performance under all scenarios.
Particularly, at peak time, where accurate failure diagnosis is critical to guarantee the user experience thus urgently required by telecom operators, TelOps demonstrates an indisputable advantage in diagnosis accuracy (15.8\%$\sim$28.0\% higher).
Random Forest performs well under both `All Day' and `Off-Peak' scenarios, whose accuracy significantly drops at peak time (at least 24.6\% lower).
CNN and AIOps perform similarly under all scenarios, where the latter achieves slightly higher diagnosis accuracy.

The performance of different comparatives is heavily influenced by real-world TN failure patterns under different scenarios.
In the off-peak scenario, most failures are commonly encountered and can be well explained by expert knowledge, where Random Forest performs well with explicit expert rules.
In the peak scenario, failures are usually more complex and rarer, and the corresponding expert knowledge is often insufficient and highly fragmented.
In this case, Random Forest cannot effectively extract latent failure patterns, leading to poor generalization performance.
Neural network-based methods (i.e., CNN and AIOps) with general model structures perform more steadily due to their relatively better latent pattern extraction.
However, their diagnosis accuracy is severely limited due to the lack of subtle knowledge embedding steps introduced by TelOps.

\section{Opportunities and Challenges to AI-driven O\&M for TNs}\label{sec:open}
We have demonstrated the effectiveness of TelOps above.
There are still several important open research issues of AI-driven O\&M for TNs deserving further elaboration.

\subsection{Systematic O\&M Knowledge Embedding}
O\&M knowledge embedding is a critical part of TelOps.
In our case study, a naive solution is applied, i.e., using the association analysis result to construct the GNN structure.
For enormous O\&M knowledge for TNs in practice, there are various possible forms to embed different knowledge, and the efficiency of our naive method would be limited.
The major challenge is to construct a systematic methodology to embed different O\&M knowledge and the corresponding restriction.
We believe that the first step is to accurately classify various O\&M knowledge, where knowledge similarity evaluation considering the form, underlay mechanism, and expert cognition is the core issue to address.

\subsection{Automatic O\&M Knowledge Reusing}
Knowledge as a Service (KaaS) is one of the cores to construct Autonomous Networks~\cite{tmf2021an}.
In particular, automatic knowledge reusing is essential to autonomous O\&M for TNs.
To achieve this, O\&M knowledge needs to be represented and managed in a unified manner that can be flexibly embedded into different O\&M solutions.
In our vision, neural networks are the desired form since they can be easily integrated with different algorithm components.

\subsection{Resource Provisioning for TNs}
The emergence of NFV, SDN, and 5G network slicing technologies implies the development of TNs to a higher virtualization level.
In this case, we believe that resource provisioning for TNs (i.e., seeking a reasonable provisioning scheme for all available resources) is a significant issue to address, especially considering the enormous resource waste caused by the current TN construction strategy.
In fact, bandwidth increasing and redundant line addition are intuitively conducted in current TNs to ensure network availability.
Moreover, since TNs are fundamental infrastructures, we believe that guaranteeing operation safety will be the primary consideration for TN resource provisioning.

\section{Conclusion}\label{sec:conclusion}
This article presents a systematic investigation on AI-driven O\&M for TNs.
Focusing on TNs' inherent properties, we propose TelOps, the first AI-driven O\&M framework for TNs.
According to the layering architecture of TelOps, we study both essential preventive and reactive O\&M tasks at the application layer, and provide systematic support from the physical to machine learning layers.
TelOps is comprehensively compared with AIOps, the popular IT system O\&M framework.
Results of the case study on the failure diagnosis for a real industrial MAN clearly demonstrate the performance gain of TelOps.
We further discuss opportunities and challenges in the future design of knowledge embedding, reusing, and resource provisioning in O\&M for TNs.

Autonomous networks are the blueprint of future TNs.
TelOps opens a new door to applying AI techniques for automatic O\&M for TNs.
We expect TelOps to inspire more insights on AI-driven TN automation.

\bibliography{ref}
\bibliographystyle{IEEEtran}

\begin{IEEEbiographynophoto}{Yuqian Yang}
received his MSc degree in Business Intelligence from Efrei Paris engineering school of digital technologies, France, in 2014.
He is pursuing his Ph.D. degree in the School of Computer Science and Technology at Xi'an Jiaotong University.
His current research interests include AIOps, knowledge-embedded data mining, and graph neural networks.
\end{IEEEbiographynophoto}

\begin{IEEEbiographynophoto}{Shusen Yang}
received his Ph.D. degree in Computing from Imperial College London in 2014.
He is a professor and deputy director of the National Engineering Laboratory for Big Data Analytics, and deputy director of Ministry of Education Key Lab for Intelligent Networks and Network Security, both at Xi'an Jiaotong University (XJTU), China.
Shusen is a DAMO Academy Young Fellow, and an honorary research fellow at Imperial College London.
He is a senior member of IEEE.
His research focuses on distributed systems and data sciences, and their applications in industrial scenarios.
\end{IEEEbiographynophoto}

\begin{IEEEbiographynophoto}{Cong Zhao}
received his Ph.D. degree in Computer Science and Technology from Xi'an Jiaotong University (XJTU) in 2017.
He worked as a research associate at Imperial College London from 2018 to 2022.
He is currently an associate professor at XJTU.
His research interests include scenario-driven distributed intelligent systems and industrial small-sample learning.
\end{IEEEbiographynophoto}

\begin{IEEEbiographynophoto}{Zongben Xu}
received his Ph.D. degree in mathematics from Xi’an Jiaotong University, China, in 1987.
He served as the Vice President of Xi’an Jiaotong University from 2003 to 2014.
He is the Chief Scientist of National Basic Research Program of China (973 Project), and the Director of the Institute for Information and System Sciences, Xi’an Jiaotong University.
His current research interests include intelligent information processing and applied mathematics.
He delivered a 45 minutes talk on the International Congress of Mathematicians 2010.
He was elected as member of Chinese Academy of Science in 2011.
\end{IEEEbiographynophoto}

\vfill

\end{document}